\documentclass[sigconf]{acmart}

\AtBeginDocument{%
  }

\copyrightyear{2026}
\acmYear{2026}
\setcopyright{cc}
\setcctype{by}
\acmConference[OASIS '26]
  {The 1st ACM SIGSPATIAL Student Challenge on Open Agents with Spatial Intelligence for Social Good}
  {November 03--06, 2026}
  {Riverside, CA, USA}
\acmBooktitle{The 1st ACM SIGSPATIAL Student Challenge on Open Agents with Spatial Intelligence for Social Good (OASIS '26), November 03--06, 2026, Riverside, CA, USA}
\acmDOI{10.1145/3849739.3856761}
\acmISBN{979-8-4007-3061-0/2026/11}

\usepackage{enumitem}
\setlist[itemize]{leftmargin=*,labelindent=0pt,labelsep=0.4em,topsep=2pt,itemsep=2pt,parsep=0pt}
\setlist[enumerate]{leftmargin=*,labelindent=0pt,labelsep=0.4em,topsep=2pt,itemsep=2pt,parsep=0pt}

\begin{document}

\title{A Topology-Guided GeoAI Agentic Framework for Urban Flood Screening}

\author{Yutian Tao, Haoan Feng, Xin Xu, and Leila De Floriani} 
\email{\{tyt16888, hfengac, xinxu629, deflo\}@umd.edu} \affiliation{        \institution{University of Maryland} \city{College Park} 
\state{Maryland} \country{USA} 
} 
\correspondingauthor

\begin{abstract}
Open-world geographic artificial intelligence (GeoAI) agents can transform public geospatial data into accessible decision support for real-world challenges, but must balance autonomous reasoning with reliable spatial analysis and human oversight. We present a topology-guided GeoAI agentic framework for urban flood screening that takes a place name, drawn region, or uploaded boundary and returns ranked candidate basins with an interactive map and report. A deterministic controller retrieves public elevation, rainfall, soil, land-cover, exposure, and infrastructure layers, extracts terrain depressions with discrete Morse theory, and ranks basins with a relative flood-priority score. The language model acts only at recorded decision points: selecting a persistence threshold from a precomputed candidate table, drafting recommendations from computed evidence, and reviewing each recommendation against the same evidence. Three human-in-the-loop checkpoints let users revise the threshold, factor weights, or recommendation. In a College Park, Maryland case study, the selected threshold reduced 278 raw terrain minima to 70 basins, with 77\% of these basins' lowest points also identified as sinks by a separately computed eight-direction (D8) flow-routing model. Across four further study areas, agreement ranged from 61\% to 83\%. By combining open data, explicit spatial reasoning, traceable evidence, and human oversight, the framework enables transparent first-pass flood screening while keeping consequential recommendations inspectable and user-controlled.
\end{abstract}



\maketitle
\thispagestyle{empty}

\section{Introduction}
Urban flooding can disrupt roads, buildings, and drainage infrastructure when heavy rainfall overwhelms local drainage capacity. A terrain-based flood assessment usually requires specialist hydrological modeling, licensed software, and data preparation that smaller municipalities, campuses, and community organizations may not be able to support \cite{Petersson2020,Pirone2026}. These institutions need a screening tool that identifies locations worth closer inspection and exposes the assumptions behind each ranking.

Terrain-based screening begins with a digital elevation model (DEM). Topological data analysis summarizes the terrain basins and drainage structure of this elevation field. In the discrete Morse analysis used here, every local minimum of the DEM except the global minimum is paired with the saddle at which its basin merges with a deeper basin \cite{forman2002user,edelsbrunner2010computational}. The elevation difference between the paired minimum and saddle is the \emph{persistence} of that basin. A \emph{persistence threshold} is the smallest persistence at which a basin is retained. Basins below the threshold are removed as noise. After simplification, the cells that drain toward each retained minimum form a \emph{Morse basin}, and we call it a \emph{retained basin} when the distinction from removed depressions matters.

Choosing the threshold is the first technical gap. A DEM of an urban area contains hundreds of shallow basins produced by measurement noise and small surface features. Because persistence depends on terrain relief, DEM resolution, and noise characteristics, no single threshold applies across study areas \cite{bobrowski2023universal}. The threshold must therefore be chosen per study area, and the choice must be recorded, as it determines every downstream count and ranking.

The second gap concerns hydrological relevance. Persistence captures topological prominence but does not ensure physically plausible runoff. We therefore use D8 flow routing as an independent check \cite{o1984extraction,w14152363}. For each retained basin, we test whether its minimum is also a \emph{D8 sink}, a cell with no downslope neighbor, and compare its \emph{Morse contributing area}, the area assigned to that basin, with the \emph{D8 flow-accumulation area}, the upstream area whose D8 flow path passes through the same minimum. Together, these checks measure consistency between two representations derived from the same DEM.

\begin{figure*}[t]
  \centering
  \includegraphics[width=\textwidth]{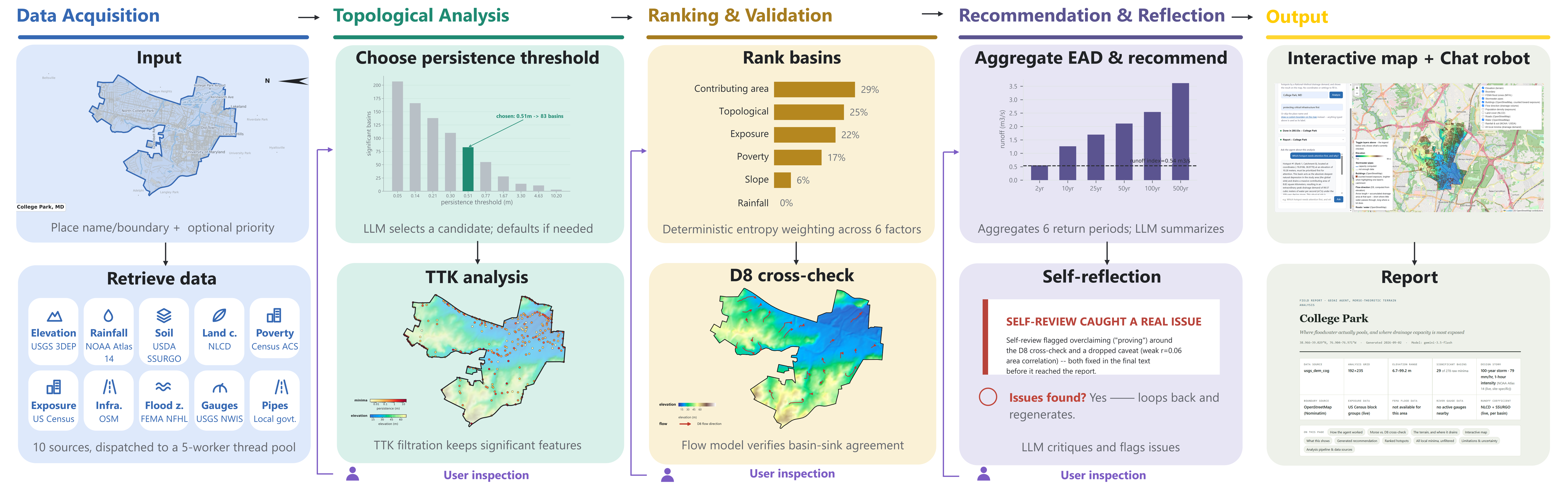}\vspace{-0.35cm}
  \caption{Workflow of the framework with intermediate results. Deterministic stages retrieve and align data, extract and rank Morse basins, and compare them with D8 routing and with re-ranking under alternative weights. Separate language-model calls select the persistence threshold, draft the recommendation, review it against the recorded evidence, and answer later user questions. Purple arrows mark the three human-in-the-loop (HITL) checkpoints.}
  \Description{A horizontal workflow with five columns. The first column shows the College Park boundary and ten retrieved data categories. The second shows a bar chart of candidate persistence thresholds against basin counts, with the selected threshold highlighted, above a terrain map with retained minima. The third shows entropy-derived factor weights above a terrain map with the D8 flow network overlaid. The fourth shows peak runoff across return periods above a callout describing a self-review that corrected an overclaiming draft. The fifth shows the interactive map, the question-answering interface, and the generated report. Purple arrows connect three checkpoints back to the threshold, ranking, and recommendation stages.}
  \label{fig:pipeline}\vspace{-0.4cm}
\end{figure*}

Existing agentic approaches (\emph{e.g.}, ReAct\cite{yao2022react}, Plan-and-Solve\cite{wang2023plan}, and Reflexion\cite{shinn2023reflexion}) allow language models to interleave reasoning, planning, and self-critique. For flood screening, however, open-ended reasoning can make numerical results depend on model outputs. We therefore keep geospatial computation deterministic and restrict the language model to recorded decision points for threshold selection, evidence-grounded recommendation, self-review, and question answering. Users can intervene at three human-in-the-loop (HITL) checkpoints. Our contributions are threefold:
\begin{itemize}
    \item A GeoAI agentic framework that combines predefined queries to public data services, deterministic terrain analysis, and language-model decisions restricted to recorded choice points.
    \item A topology-guided screening method that selects a persistence threshold from a candidate table, ranks basins with a relative flood-priority score, and checks the result against a D8 flow-routing model of the same DEM.
    \item An evaluation across five study areas that reports threshold selection, D8 agreement, first-rank stability under four weighting schemes, and the behavior of self-review and user feedback.
\end{itemize}

\section{GeoAI Agentic Framework}
\label{sec:framework}

\subsection{System Overview and Data Acquisition}

Figure~\ref{fig:pipeline} summarizes the framework. A run starts from a place name, user-drawn polygon, or uploaded boundary, with optional priorities, ranking weights, and design-storm return period (default: 100 years). The framework retrieves and aligns public data, extracts and ranks terrain basins, performs consistency checks, and generates an evidence-grounded report. The language model is invoked only at designated decision points, while computed quantities and intermediate results are stored in an \emph{evidence bundle}. For fixed inputs, numerical stages are deterministic.

Data are retrieved through predefined adapters to the public sources listed in Table~\ref{tab:data-contract}. Elevation is required; missing optional layers are omitted or replaced by documented defaults. All layers are aligned to the elevation grid, and substitutions are recorded in the evidence bundle and report.

\begin{table*}[t]
\centering
\caption{Public data sources used by the framework.}
\label{tab:data-contract}
\vspace{-3mm}
\begin{tabular}{p{0.25\textwidth} p{0.35\textwidth} p{0.35\textwidth}}
\toprule
\textbf{Data category} & \textbf{U.S. source} & \textbf{Non-U.S. source} \\
\midrule
Elevation & USGS 3DEP & Copernicus GLO-30 \\
Land cover & NLCD & Sentinel-2 10m \\
Exposure & Census block-group population density & WorldPop 100m \\
Infrastructure & OSM roads, buildings, water & OSM roads, buildings, water \\
Rainfall & NOAA Atlas 14 & N/A \\
Soil & USDA SSURGO & N/A \\
Socioeconomic vulnerability & Census-tract poverty rate (ACS) & N/A \\
Stormwater pipes & Local government data & N/A \\
Flood zones & FEMA flood-hazard data & N/A \\
Stream gauges & USGS stream-gauge service & N/A \\
\bottomrule
\end{tabular}
\vspace{-4mm}
\end{table*}

\subsection{Topology-Guided Basin Extraction}
\label{sec:topology}

The framework computes terrain topology using the Topology ToolKit (TTK)~\cite{tierny2017topology}. A preliminary pass collects the persistence of finite-persistence minima in the DEM. The controller then constructs candidate persistence thresholds from the 25th, 40th, 50th, 60th, 70th, 80th, 90th, 95th, and 99th percentiles of this distribution, together with a default equal to 5\% of the DEM elevation range. Each candidate is associated with the number of minima it would retain.

Threshold selection is the first language-model decision. Given the candidate table, study-area name, and user-specified priority, the model selects one candidate and justifies the choice using the corresponding change in basin count. The controller then performs a second TTK pass at the selected threshold to extract the retained basins. Both the candidate-based estimate and the achieved basin count are recorded.

\subsection{Basin Prioritization}
\label{sec:ranking}

To rank the retained basins by flood priority, each retained basin $b$ is characterized by six normalized factors: topological prominence, contributing area, rainfall intensity, inverse local slope, exposure, and socioeconomic vulnerability. The Morse contributing area $A_b$ is the number of cells assigned to the basin multiplied by cell area. Topological prominence is derived from minimum persistence, contributing area is normalized by the largest retained basin, rainfall intensity by the maximum value in the design-storm table, and exposure by the range of the exposure raster. Inverse slope gives higher values to flatter basin floors. Missing slope, exposure, or poverty-rate values receive 0.5, representing a neutral value that avoids treating missing data as either low or high risk.

Given nonnegative weights $w_j$ summing to one, the relative flood-priority score is
\begin{equation}
P_b = \sum_{j \in \{t,a,r,s,e,v\}} w_j x_{b,j}.
\label{eq:priority}
\end{equation}
The score provides a within-run ranking rather than a calibrated flood probability or inundation estimate. By default, weights are derived using the entropy weight method~\cite{zhu2020effectiveness}, which assigns greater weight to factors with greater variation across retained basins. Because rainfall is spatially uniform within a run, it receives zero entropy-derived weight. Users may instead select a fixed weight preset or provide custom weights. A weight-sensitivity check re-scores the same retained basins under three fixed weighting presets (balanced, life-safety, infrastructure-protection) and the entropy weights above, then reports whether the top-ranked basin is stable across all four schemes.

\subsection{Screening-Level Runoff and Consistency Checks}
\label{sec:runoff}
The framework additionally estimates screening-level peak runoff to support recommendation generation, while this quantity is not included in the priority score in Equation~\ref{eq:priority}. For each basin, peak discharge under return period $T$ is estimated using the Rational Method~\cite{kuichling1889relation} as $Q_b(T)=C_b I(T)A_b/3.6$ in m$^3$/s, where $T$ is the storm return period in years, $I(T)$ is the corresponding design-storm rainfall intensity in mm/h, $A_b$ is the basin's contributing area in km$^2$, and $C_b$ is a dimensionless runoff coefficient derived from soil and land cover using the SCS Curve Number procedure~\cite{USDA-TR55}. The framework integrates $Q_b(T)$ across the 2-, 10-, 25-, 50-, 100-, and 500-year scenarios using the trapezoidal rule with AEP, $1/T$, as the integration variable, following the approach used in the expected annual damage framework~\cite{USACE-EAD}. The resulting metric is provided as supporting evidence for recommendation generation rather than as a component of the flood-priority score.

For consistency checking, the controller computes D8 flow routing on the same elevation grid. It compares whether retained Morse minima coincide with D8 sinks and compares Morse contributing area with D8 flow-accumulation area at the corresponding minimum cells. The first comparison is summarized by the D8 sink-agreement rate; the second uses Pearson correlation and the coefficient of determination ($R^2$) after log transformation. These checks provide independent diagnostics of the topology-derived basins and are also performed across additional study areas.

\subsection{Recommendation and Human Oversight}
\label{sec:interaction}

Recommendation generation and evidence-grounded self-review are separate language-model calls. The drafting call receives the evidence bundle and user priority, while the review call checks numerical consistency, unsupported certainty, and omitted data limitations. Detected problems trigger either textual correction or a bounded re-analysis of threshold selection or basin ranking, followed by regeneration and review.

Human oversight is provided through three checkpoints covering threshold selection, basin ranking, and recommendation text. Users can revise the selected threshold, factor weights, or recommendation, respectively. 
Language-model failures do not modify completed numerical results: threshold selection falls back to the default candidate, failed reviews are marked as unreviewed, and unavailable analytical checks are explicitly reported.

\section{Evaluation}
\label{sec:evaluation}
We evaluate the framework along three dimensions: terrain simplification, cross-model spatial consistency, and robustness of the final ranking and recommendation. College Park, Maryland serves as the end-to-end case study, while spatial consistency is additionally evaluated in four study areas: Annapolis and Baltimore, Maryland, and Richmond and Alexandria, Virginia. The language model calls used Google Gemini-3.5-flash \cite{gemini35flash}.

The College Park run used the user-specified priority "protecting the university campus and nearby residential streets from flooding." Ten data categories returned data except stream gauges: a 192×235 USGS 3DEP elevation grid, NLCD land cover, 26 FEMA flood-zone features, a 100-year rainfall intensity of 79 mm/h, hydrologic soil group C, US Census block-group population density, US Census-tract poverty rate, 5,936 stormwater pipe segments, 12,138 roads, 20,948 buildings, and 291 water features. The USGS gauge query returned no active gauge within the study area, which the report records as a verified empty result.

The first TTK pass identified 278 raw local minima. Ten candidates were evaluated, spanning the 25th--99th persistence percentiles plus a default threshold of 5\% of the elevation range, with estimated basin counts ranging from 207 to 3. The language model selected the 70th-percentile candidate, 0.508~m, with an estimated 83 retained minima. Its justification cited the drop in basin count from 83 to 55 at the next candidate (0.774~m), describing 0.508~m as capturing meaningful depressions associated with surface ponding while filtering out minor surface irregularities.

\begin{figure}[t]
    \centering
    \includegraphics[width=\columnwidth]{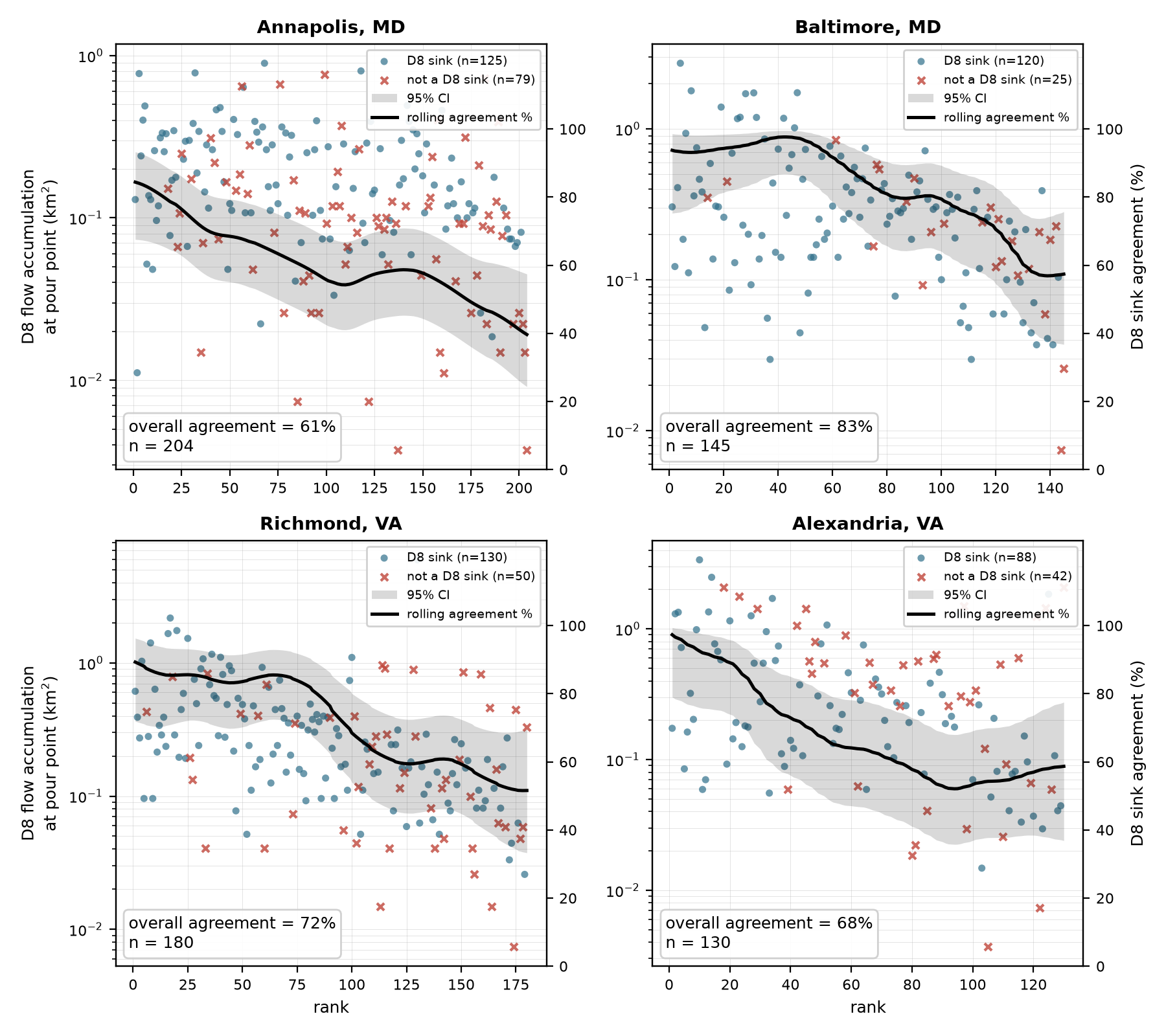}\vspace{-0.4cm}
    \caption{D8 sink status of retained basin minima in four study areas. Each point is one retained basin, ordered by its rank under entropy-derived weights. The vertical axis gives the D8 flow-accumulation area at the basin minimum on a logarithmic scale. Each panel reports the D8 sink-agreement rate $A_{\mathrm{sink}}$ (the fraction of that area's retained basins whose minimum coincides with a D8 sink) and the number of retained basins $n$.}
    \Description{Four scatter panels, one per study area. Points are ordered by basin priority rank. Blue circles indicate minimum cells that are D8 sinks, and red crosses indicate cells that are not. The vertical axis shows D8 flow-accumulation area on a logarithmic scale.}
    \label{fig:sink}\vspace{-0.4cm}
\end{figure}

\begin{figure}[t]
    \centering
    \includegraphics[width=\columnwidth]{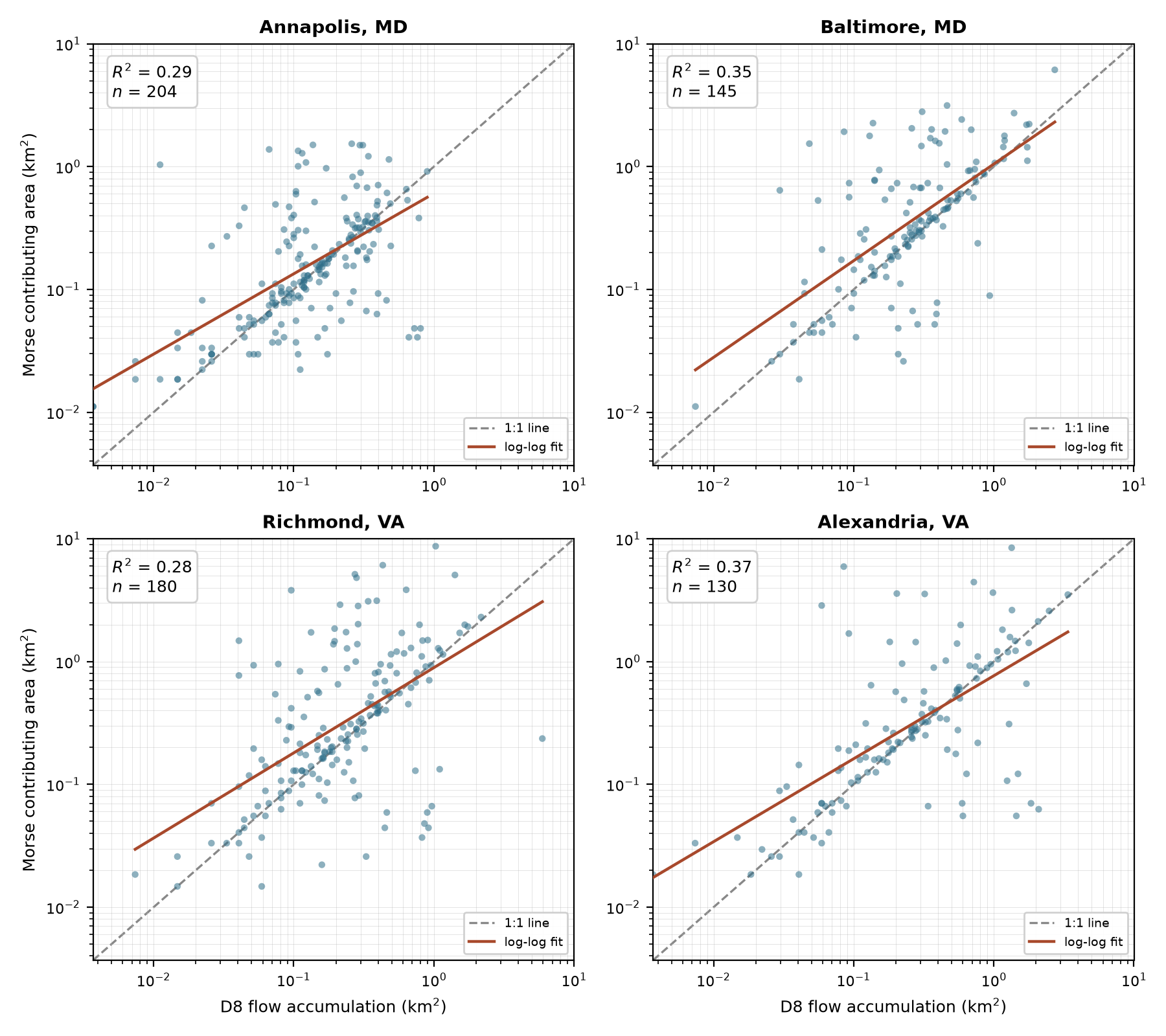}\vspace{-0.4cm}
    \caption{D8 flow-accumulation area (horizontal) against Morse contributing area (vertical) for the retained basins of Figure~\ref{fig:sink}, both on logarithmic scales, with a linear fit in log space, $r_{\log}$, $R^2$, and $n$ per panel.}
    \Description{Four log-log scatter plots comparing D8 flow-accumulation area on the horizontal axis with Morse contributing area on the vertical axis. Each panel contains blue points, a fitted red line, and annotations for correlation, coefficient of determination, and sample size.}
    \label{fig:area}
\end{figure}

D8 routing on the same grid identified 222 raw sinks, compared with 278 raw Morse minima. Of the 70 retained basins, 77\% had a minimum cell that was also a D8 sink. The minimum cells of the remaining 23\% of basins lie mostly on flat areas, where D8 assigns a downslope neighbor to a cell that Morse analysis retains as a minimum. 
The same D8 consistency checks were also performed across the additional study areas. Figure~\ref{fig:sink} summarizes sink agreement across these areas, while Figure~\ref{fig:area} compares the D8 flow-accumulation area with the Morse contributing area.

Under entropy weighting, catchment 8 (38.977\textdegree{}N, 76.919\textdegree{}W) ranked first, with a contributing area of 0.249~km$^2$ and a relative flood-priority score of 0.61. The basin lies within FEMA's Special Flood Hazard Area and 6.8~m from the nearest mapped water feature, providing independent spatial evidence consistent with the topological ranking. It remained first under all four weighting schemes tested: balanced, life-safety, infrastructure-protection, and entropy.

\section{Conclusion}
We presented a topology-guided GeoAI agentic framework for urban flood screening that integrates public geospatial data, deterministic spatial analysis, LLM-based reasoning, and human-in-the-loop review. Across five study areas, the framework showed consistent agreement between Morse-based terrain structures and independent D8 routing, while sensitivity analysis and self-review helped identify unstable rankings and unsupported statements.

Beyond the technical evaluation, the framework is designed to make preliminary flood-risk screening more accessible and transparent. It provides a low-cost first-pass analysis from public data, keeps recommendations traceable to their underlying spatial evidence, and allows users to inspect results through an interactive map, report, and LLM interface. Human checkpoints further ensure that consequential decisions remain reviewable rather than being delegated entirely to the language model.

The experiments also highlight important lessons and limitations. LLM-generated recommendations can overstate available evidence, reinforcing the need for explicit verification and human oversight. The current evaluation uses a limited number of study areas, U.S.-specific data services, simplified exposure representation, and no observed flood events as ground truth. Future work will therefore evaluate more geographic settings, incorporate broader and more complete data sources, compare multiple LLMs, and further assess the framework against observed or simulated flood information.

\bibliographystyle{ACM-Reference-Format}
\bibliography{references}

@article{Petersson2020,
  author  = {Petersson, Louise and ten Veldhuis, Marie-Claire and Verhoeven, Govert and Kapelan, Zoran and Maholi, Innocent and Winsemius, Hessel C.},
  title   = {Community Mapping Supports Comprehensive Urban Flood Modeling for Flood Risk Management in a Data-Scarce Environment},
  journal = {Front. Earth Sci.},
  volume  = {8},
  year    = {2020},
  doi     = {10.3389/feart.2020.00304}
}

@article{Pirone2026,
  author  = {Pirone, Dina and Padulano, Roberta and Del Giudice, Giuseppe},
  title   = {Approaching Urban Flood Modelling According to Available Data: A State-of-the-Art Review and Future Challenges},
  journal = {J. Flood Risk Manag.},
  volume  = {19},
  number  = {1},
  pages   = {e70184},
  year    = {2026},
  doi     = {10.1111/jfr3.70184}
}

@book{edelsbrunner2010computational,
  author    = {Edelsbrunner, Herbert and Harer, John},
  title     = {Computational Topology: An Introduction},
  year      = {2010},
  publisher = {American Mathematical Society}
}

@article{forman2002user,
  author  = {Forman, Robin},
  title   = {A User's Guide to Discrete Morse Theory},
  journal = {S{\'e}m. Lothar. Combin.},
  volume  = {48},
  pages   = {B48c},
  year    = {2002}
}

@article{bobrowski2023universal,
  author  = {Bobrowski, Omer and Skraba, Primoz},
  title   = {A Universal Null-Distribution for Topological Data Analysis},
  journal = {Sci. Rep.},
  volume  = {13},
  pages   = {12274},
  year    = {2023}
}

@article{tierny2017topology,
  author  = {Tierny, Julien and Favelier, Guillaume and Levine, Joshua A. and Gueunet, Charles and Michaux, Michael},
  title   = {The Topology ToolKit},
  journal = {IEEE Trans. Vis. Comput. Graph.},
  volume  = {24},
  number  = {1},
  pages   = {832--842},
  year    = {2017}
}

@article{o1984extraction,
  author  = {O'Callaghan, John F. and Mark, David M.},
  title   = {The Extraction of Drainage Networks from Digital Elevation Data},
  journal = {Comput. Vis. Graph. Image Process.},
  volume  = {28},
  number  = {3},
  pages   = {323--344},
  year    = {1984}
}

@article{w14152363,
  author  = {D{\'a}vila-Hern{\'a}ndez, Sandra and Gonz{\'a}lez-Trinidad, Juli{\'a}n and J{\'u}nez-Ferreira, Hugo Enrique and Bautista-Capetillo, Carlos Francisco and Morales de {\'A}vila, Heriberto and C{\'a}zares Escare{\~n}o, Juana and Ortiz-Letechipia, Jennifer and Robles Rovelo, Cruz Octavio and L{\'o}pez-Baltazar, Enrique A.},
  title   = {Effects of the Digital Elevation Model and Hydrological Processing Algorithms on the Geomorphological Parameterization},
  journal = {Water},
  volume  = {14},
  number  = {15},
  pages   = {2363},
  year    = {2022},
  doi     = {10.3390/w14152363}
}

@article{yao2022react,
  author  = {Yao, Shunyu and Zhao, Jeffrey and Yu, Dian and Du, Nan and Shafran, Izhak and Narasimhan, Karthik and Cao, Yuan},
  title   = {{ReAct}: Synergizing Reasoning and Acting in Language Models},
  journal = {arXiv:2210.03629},
  year    = {2022}
}

@article{shinn2023reflexion,
  author  = {Shinn, Noah and Cassano, Federico and Berman, Edward and Gopinath, Ashwin and Narasimhan, Karthik and Yao, Shunyu},
  title   = {Reflexion: Language Agents with Verbal Reinforcement Learning},
  journal = {arXiv:2303.11366},
  year    = {2023}
}

@inproceedings{wang2023plan,
  author    = {Wang, Lei and Xu, Wanyu and Lan, Yihuai and Hu, Zhiqiang and Lan, Yunshi and Lee, Roy Ka-Wei and Lim, Ee-Peng},
  title     = {Plan-and-Solve Prompting: Improving Zero-Shot Chain-of-Thought Reasoning by Large Language Models},
  booktitle = {Proc. 61st Annu. Meeting Assoc. Comput. Linguist. (ACL)},
  pages     = {2609--2634},
  year      = {2023}
}

@article{zhu2020effectiveness,
  author  = {Zhu, Yuxin and Tian, Dazuo and Yan, Feng},
  title   = {Effectiveness of Entropy Weight Method in Decision-Making},
  journal = {Math. Probl. Eng.},
  volume  = {2020},
  pages   = {3564835},
  year    = {2020}
}

@article{kuichling1889relation,
  author  = {Kuichling, Emil},
  title   = {The Relation between the Rainfall and the Discharge of Sewers in Populous Districts},
  journal = {Trans. ASCE},
  volume  = {20},
  pages   = {1--56},
  year    = {1889}
}

@techreport{USDA-TR55,
  author      = {{USDA Natural Resources Conservation Service}},
  title       = {Urban Hydrology for Small Watersheds},
  institution = {USDA NRCS},
  number      = {TR-55},
  year        = {1986}
}

@techreport{USACE-EAD,
  author      = {{U.S. Army Corps of Engineers}},
  title       = {Risk-Based Analysis for Flood Damage Reduction Studies},
  institution = {USACE},
  number      = {EM 1110-2-1619},
  year        = {1996}
}

@misc{gemini35flash,
  author       = {{Google DeepMind}},
  title        = {{Gemini 3.5 Flash Model Card}},
  year         = {2026},
  howpublished = {\url{https://deepmind.google/models/model-cards/gemini-3-5-flash/}},
  note         = {Accessed: 2026-09-29}
}

\end{document}